\documentclass[accepted]{ecai} 

\usepackage{latexsym}
\usepackage{amssymb}
\usepackage{amsmath}
\usepackage{amsthm}
\usepackage{booktabs}
\usepackage{enumitem}
\usepackage{graphicx}
\usepackage{natbib} 
    \usepackage{color}

\newcommand{\BibTeX}{B\kern-.05em{\sc i\kern-.025em b}\kern-.08em\TeX}

\title{Towards Robust Continual Learning With Bayesian Adaptive Moment Regularization}

\author[1,2]{\href{mailto:<jwf40@cam.ac.uk>?Subject=BAdam Paper}{Jack Foster}{}}
\author[1,2]{Alexandra Brintrup}
\affil[1]{%
    Dept. Engineering\\
    University of Cambridge\\
    Cambridge, UK
}
\affil[2]{%
The Alan Turing Institute\\
London, UK
}

\begin{document}
\maketitle



\begin{abstract}
The pursuit of long-term autonomy mandates that machine learning models must continuously adapt to their changing environments and learn to solve new tasks. Continual learning seeks to overcome the challenge of catastrophic forgetting, where learning to solve new tasks causes a model to forget previously learnt information. Prior-based continual learning methods are appealing as they are  computationally efficient and do not require auxiliary models or data storage. However, prior-based approaches typically fail on important benchmarks and are thus limited in their potential applications compared to their memory-based counterparts. We introduce Bayesian adaptive moment regularization (BAdam), a novel prior-based method that better constrains parameter growth, reducing catastrophic forgetting. Our method boasts a range of desirable properties such as being lightweight and task label-free, converging quickly, and offering calibrated uncertainty that is important for safe real-world deployment. Results show that BAdam achieves state-of-the-art performance for prior-based methods on challenging single-headed class-incremental experiments such as Split MNIST and Split FashionMNIST, and does so without relying on task labels or discrete task boundaries.
\end{abstract}


%
%


\section{Introduction}\label{introduction}

A common assumption in machine learning (ML) is that training data is independent and identically distributed (i.i.d). However, in continual learning (CL) an ML model is presented with a non-i.i.d stream of sequential tasks. Under such circumstances traditional learning methods suffer from a phenomenon known as catastrophic forgetting, where previously learnt knowledge is lost as a result of model parameters biasing towards recently observed data \citep{mccloskey1989catastrophic, THRUN199525, french1999catastrophic}. Continual learning is a major, long-standing challenge within machine learning \citep{hassabis2017neuroscience}. A canonical example of continual learning is that of a Mars rover needing to adapt to ever changing terrain as it traverses the surface \citep{farquhar2018towards}.\\

There are several approaches to continual learning, such as memory-based, where past experiences are stored in a replay-buffer, and (prior-based) regularization methods, which seek to constrain weight updates based on their importance to previously learned tasks. Regularization methods are attractive as they are biologically accurate \citep{hassabis2017neuroscience}, do not require additional memory storage, and typically do not grow in computational complexity as the number of tasks increases. These properties make regularization approaches particularly appealing for applications that must operate in real-time, in the real-world, or under strict hardware constraints.\\

In \citet{farquhar2018towards}, five desiderata for CL evaluations are introduced which ensure that evaluations are robust and representative of real-world CL challenges. They are as follows: 
\begin{itemize}
    \item Input data from later tasks must bear some resemblance to previous tasks
    \item Task labels should not be available at test-time (since if they were then separate models could be trained for each task)
    \item All tasks should share an output head (multi-head architectures simplify the problem by making the assumption that the task label is known at test-time)
    \item There cannot be unconstrained retraining on old tasks (motivations of CL preclude this)
    \item  CL evaluations must use more than two tasks
\end{itemize} 

Two experiments that satisfy these requirements are the single-headed Split MNIST \citep{pmlr-v70-zenke17a}, and Split FashionMNIST problems. Prior-based methods typically fail to improve upon na\"ive baseline performance in these evaluations, experiencing the same catastrophic forgetting as ordinary supervised learning. These methods require the experimental setup to violate the desiderata in order to perform well (e.g. using multiple heads and test-time task labels). This is a significant barrier to the application of prior-based methods.\\

A further weakness of regularization approaches is they often make strong assumptions about the structure of the data stream. Specifically, it is typically assumed that tasks are discrete (i.e. no overlaps), and that the boundaries between tasks are known. In \citet{van2019three} a different paradigm is noted, where task boundaries are unknown, and where tasks may overlap. This is a much harder problem as task labels cannot be used, and also because taking multiple passes over each task's data is no longer possible. While such experimental conditions are very challenging, they may be more representative of real-world online continual learning challenges, where there is no known structure to the data stream. Bayesian Gradient Descent (BGD) was proposed as a solution to continual learning with unknown, graduated task boundaries by solving the problem with a closed-form update rule that does not rely on task labels \citep{zeno2018task}. BGD is insightful and possesses several appealing properties. BGD requires no auxiliary memory storage, it does not require task labels, it does not require tasks to be separated discretely, and finally it has calibrated uncertainties which is valuable in safety-critical applications such as robotics \citep{jospin2022hands}. However, BGD has two key failings: first it is slow to converge and second, BGD fails to solve single-headed class-incremental problems.\\

We present Bayesian Adaptive Moment Regularization (BAdam), a novel continual learning method that unifies the closed-form update rule of BGD with properties found in the Adam optimizer \citep{kingma2014adam}, maintaining adaptive per-parameter learning rates based on the first and second order moment of the gradient. The consequence of this is improved convergence rates, more effective constraining of parameter updates and therefore, we hypothesize, the learning of parameters that better generalize across multiple tasks. We show empirically that our method makes substantial improvements over previous prior-based approaches on single-headed class-incremental problems.\\ 

Finally, there has been no evaluation of methods in an environment that enforces the desiderata from \citet{farquhar2018towards}, while also featuring graduated task boundaries (where tasks have some overlap as one finishes and another begins), no train-time task labels, and that permits only a single-epoch of training per task. This formulation is more reflective of the scenarios encountered in the real-world, where the environment often changes gradually, and retraining over multiple epochs is hard due to compute restrictions. Therefore, in this work we introduce this formulation of a continual learning challenge and evaluate regularization methods in such an environment.\\

To summarise, our contributions are as follows:
\begin{itemize}
    \item A novel continual learning method that achieves state-of-the-art performance compared to previous prior-based approaches. This is achieved by improving the protective properties of BGD through more efficient estimation of the parameter mean, thereby yielding more effective uncertainty estimates.
    \item The introduction of a new evaluation environment that is more reflective of real-world scenarios, featuring no task labels, a single epoch of training, and graduated task boundaries.
\end{itemize}

\section{Related Work}\label{related}

As discussed in \citet{van2019three}, there are three primary continual learning scenarios; task-, domain-, and class-incremental learning, differentiated by task label availability (Table \ref{3ScenariosTable}). Task-IL problems are the easiest, as the label is given and thus task-specific components, such as separate heads or sub-networks, may be utilized. Such scenarios are less likely to be present in real-world tasks, due to the uncertainty surrounding the observed datastream. Class-IL scenarios are the most challenging, and class-incremental split MNIST is the simplest benchmark that accurately evaluates the efficacy of CL methods \citep{farquhar2018towards}. There exist many approaches to solving CL problems, which can be loosely classified into the following four paradigms: memory-based \citep{rolnick2019experience, lopez2017gradient, castro2018end, titsias2019functional, aljundi2019gradient, rebuffi2017icarl}, architectural \citep{rusu2016progressive, terekhov2015knowledge, draelos2017neurogenesis},  data-augmentation  \citep{li2017learning, shin2017continual, rannen2017encoder, parisi2019continual}, prompt-based \cite{wang2022learning, wang2022dualprompt}, and regularization. Here, we restrict our review to regularization methods.\\ 
\begin{table}[!tpb]
\centering
\caption{Overview of the Three Continual Learning Scenarios From\citet{van2019three}}\label{3ScenariosTable}
\begin{center}
\begin{tabular}{|c|c|}
\hline
\textbf{Scenario} & \textbf{Task-ID}\\
\hline
\hline
Task-IL & Task-ID provided\\
\hline
Domain-IL & Task-ID not provided\\
\hline
Class-IL & Must infer task-ID\\
\hline
\end{tabular}
\end{center}
\end{table}

Regularization techniques are an abstraction of biological neuro-plasticity \citep{cohen1997functional, cichon2015branch}, reducing parameter update rates proportional to their importance to previous tasks, thereby protecting past knowledge. This is effective as many parameter sets will yield equivalent performance for a given neural network \citep{hecht1992theory, sussmann1992uniqueness}, thus constraining parameter updates does not preclude strong solutions being found.\\

Elastic weight consolidation (EWC) protects knowledge of previous tasks with a quadratic penalty on the difference between the previous task's parameters, and that of the current task  \citep{kirkpatrick2017overcoming}. The Fisher information matrix of previous tasks is used to estimate parameter importance. Memory aware synapses (MAS) quantify parameter importance through measuring the sensitivity of the output to perturbations of the weight \citep{aljundi2018memory}, while synaptic intelligence (SI) calculates importance as a function of a parameter's contribution to the path integral of the gradient vector field along the parameter trajectory \citep{pmlr-v70-zenke17a}. Variational continual learning (VCL) takes a Bayesian approach \citep{nguyen2017variational}, where online variational inference is leveraged to mitigate catastrophic forgetting. A limitation of all of these methods is that they all assume task boundaries are discrete, and that the current task is known. Task-free continual learning (TFCL)  removes the reliance on task labels \citep{Aljundi_2019_CVPR}. Utilising MAS as the underlying CL method, TFCL detects plateaus in the loss surface (indicating learnt tasks), taking subsequent peaks to indicate the introduction of new tasks. This provides an online, label-free means of determining when to update importance weights, however it is somewhat fallible to graduated task boundaries, as the intermingling of tasks may cause sporadic changes in the loss surface.\\

Bayesian Gradient Descent addresses both graduated task boundaries and no task labels through a closed form update rule derived from online variational Bayes \citep{zeno2018task}. BGD's theoretical derivation assumes that data arrives in a sequential, online manner, offering a promising route to truly online continual learning. In practice, however, convergence is slow, and many epochs are required to reach optimal performance, violating the conditions described in \citet{van2019three}. Like other prior-based methods, BGD fails to solve single-headed class-incremental problems. Finally, \cite{ebrahimi2019uncertainty} propose a similar method, that also uses parameter uncertainty as an inverse measure of parameter importance. \\

Other noteworthy papers include \citet{wolczyk2022continual}, which is not a prior-based method but is a form of hard-regularization, constraining the parameter space to be within a specific hyper-rectangle. \citet{khan2018fast} Introduces a variational form of the Adam optimizer \citep{kingma2014adam}, which bares some resemblance to our proposed method, however it is not a continual learning algorithm, nor does it prevent catastrophic forgetting.

\section{Methods}\label{methods}
\subsection{Preliminaries}
\subsubsection{Problem Definition}
We consider a discriminative model trained on the dataset $\mathcal{D} = \{\mathbf{x}, \mathbf{y}\}$, where $\mathbf{x}, \mathbf{y}$ are model inputs and targets, respectfully. In a continual learning setting, $\mathcal{D}$ is comprised of $N$ distinct datasets, $\mathcal{D} = \{\mathcal{D}_{1}, ..., \mathcal{D}_{n}\}$, that the model must learn sequentially. Each dataset may be comprised of an arbitrary number of samples, including just a single datum. Since sequential learning violates the i.i.d assumption of traditional supervised learning, additional mechanisms are required to protect knowledge learned from early datasets and to prevent overfitting to recent datasets. 

\subsubsection{Bayesian Gradient Descent}
To mitigate the forgetting of prior information, BGD formulates the sequential learning problem as the task of learning the online posterior distribution of model parameters given by:
\begin{equation}
    p(\theta | \mathcal{D}_{n}) = \frac{p(\mathcal{D}_{n}|\theta)p(\theta|\mathcal{D}_{n-1})}{p(\mathcal{D}_{n})}
\end{equation}
Where $\theta$ are the model parameters, and $\mathcal{D}_{n}$ is the $n$th dataset (task) in the sequence \cite{zeno2018task}. The prior term $p(\theta | \mathcal{D}_{n-1})$ acts as a regularization term that encourages the learning of parameters for $\mathcal{D}_{n}$ that are congruent with parameters learned for $\mathcal{D}_{(n-1, ..., 1)}$. However, as calculating the exact posterior is intractable, online variational Bayes is used to approximate it. From \citet{NIPS2011_7eb3c8be}, a parametric distribution $ q(\theta | \phi) $ is used to approximate a posterior distribution by minimising the Kullback-Leibler divergence:
\begin{equation}
    KL(q(\theta | \phi) || p(\theta|\mathcal{D})) = - \mathbb{E}_{\theta \sim q (\theta|\phi)} \left[ \text{log} \frac{p(\theta|\mathcal{D})}{q(\theta|\phi)}\right]
\end{equation}

In online variational Bayes \citep{broderick2013streaming}, the optimal variational parameters are calculated by:
\begin{equation*}
    \phi^{*} = \text{arg} \min_{\phi} \int q_{n} (\theta|\phi)\text{log}\frac{ q_{n} (\theta|\phi)}{ p(\theta|\mathcal{D}_{n})} d\theta
\end{equation*}
\begin{equation*}
    \phi^{*} = \text{arg} \min_{\phi} \mathbb{E}_{\theta \sim q_{n}(\theta|\phi)} [\text{log}(q_{n}(\theta|\phi)) 
\end{equation*}
\begin{equation}\label{variation_opt}
        - \text{log}(q_{n-1}(\theta)) - \text{log}(p(\mathcal{D}_{n}|\theta)] 
\end{equation}

The following transformation is defined to arrive at our Bayesian neural network:
\begin{equation*}
    \theta_{i} = \mu_{i} + \epsilon_{i} \sigma_{i}
\end{equation*}
\begin{equation*}
    \epsilon_{i} \sim \mathcal{N}(0,1)
\end{equation*}
\begin{equation}
    \phi = (\mu, \sigma)
\end{equation}

Assuming the parametric distribution is Gaussian and that components are independent (mean-field approximation), the optimization problem in \ref{variation_opt} can be solved using unbiased Monte Carlo gradients as in \citet{blundell2015weight}. This results in the following closed form updates for the variational parameters $(\mu, \sigma)$:

\begin{equation}\label{mu_bgd}
    \mu_{t} = \mu_{t-1} - \eta \sigma_{t-1}^{2}\mathbb{E}_{\epsilon}\left[\frac{\partial(- \text{log}(p(\mathcal{D}_{n}|\theta)))}{\partial\theta_{t}}\right]
\end{equation}

\begin{equation*}\centering
    \sigma_{t} = \sigma_{t-1}\sqrt{1 + \left(\frac{1}{2}\sigma_{t-1}\mathbb{E}_{\epsilon}\left[\frac{\partial(- \text{log}(p(\mathcal{D}_{n}|\theta)))}{\partial\theta_{t}}\epsilon_{t}\right]\right)^{2}}     
\end{equation*}
\begin{equation}\label{std_dev}
    - \frac{1}{2} \sigma_{t-1}^{2} \mathbb{E}_{\epsilon} \left[\frac{\partial(- \text{log}(p(\mathcal{D}_{n}|\theta)))}{\partial\theta_{t}}\epsilon_{t}\right]
\end{equation}

Where $\eta$ is a learning rate. Since $\frac{\partial(- \text{log}(p(\mathcal{D}_{n}|\theta)))}{\partial\theta_{i}}$ depends on $\mu_{i}$ and $\sigma_{i}$, the solution is predicted by evaluating this derivative using the prior parameters. Finally, the expectations are estimated using the Monte Carlo method. For a more complete derivation of the method, we direct the reader to the original paper \citep{zeno2018task}. \\

\subsection{The BAdam Optimizer}

The update rule for $\mu$ in BGD closely resembles the update rule in stochastic gradient descent with the key difference being the variance weighting, which leads to smaller updates for highly certain parameters. This is the mechanism through which BGD lowers plasticity and prevents forgetting. While theoretically sound, in practice BGD exhibits significant forgetting in more challenging domains such as single-headed class-incremental tasks. We postulate that a contributing factor to forgetting is that slow convergence leads to large $\sigma$ values for much of training, preserving model plasticity and thereby allowing the model to forget prior knowledge. To alleviate this, we propose a new update rule for $\mu$ which leads to significantly less plasticity. First, we introduce this rule in equation \ref{badam_mu}, before exploring the intuition behind why BAdam can better protect prior knowledge:

\begin{equation*}
    m_{t} = \beta_{1} \cdot m_{t-1} + (1-\beta_{1}) \cdot \mathbb{E}_{\epsilon}\left[\frac{\partial L_{n}(\theta)}{\partial\theta_{i}}\right]
\end{equation*}

\begin{equation*}
    v_{t} = \beta_{2} \cdot v_{t-1} + (1-\beta_{2}) \cdot \left(\mathbb{E}_{\epsilon}\left[\frac{\partial L_{n}(\theta)}{\partial\theta_{i}}\right]\right)^{2}
\end{equation*}

\begin{equation*}
    \hat{m}_{t} = \frac{m_{t}}{1-\beta^{t}_{1}}
\end{equation*}

\begin{equation*}
    \hat{v}_{t} = \frac{v_{t}}{1-\beta^{t}_{2}}
\end{equation*}

\begin{equation}\label{badam_mu}
    \mu_{t} = \mu_{t-1} - \eta\sigma^{2} \cdot \frac{\hat{m}_{t}}{\sqrt{\hat{v}_{t}} + \gamma}
\end{equation}

where $L_{n}(\theta)=(- \text{log}(p(\mathcal{D}_{n}|\theta)))$ is the log likelihood loss, and $\gamma$ is a small value, typically $1e-8$. This is derived from the update rule in Adam \citep{kingma2014adam}, with $m_{t}$ being the bias-corrected mean of the loss and $v_{t}$ being the bias-corrected variance. The approach introduces a per-parameter learning rate and a  momentum characteristic, which facilitates a greater robustness to saddle points and areas of small curvature in the loss surface, as well as faster convergence \citep{kingma2014adam}.\\

Central to our hypothesis that equation \ref{badam_mu} reduces catastrophic forgetting is the understanding that in BGD a parameter's plasticity is regulated by its variance ($\sigma^2$). Minimising $\sigma$ protects knowledge, but doing so requires optimizing $\mu$. In other words, better optimization of $\mu$ leads to greater regularization for those parameters, and thus stronger protection of prior knowledge.\\


To further understand how introducing aspects of Adam into BGD can increase regularization,  we analyse discussion from \citet{zeno2018task}. Using a Taylor expansion we can see that $\mathbb{E}_{\epsilon}\left[\frac{\partial L_{n}(\theta)}{\partial\theta_{i}}\epsilon_{i}\right]$ closely approximates the curvature of the loss, for small values of $\sigma$:

   \begin{figure*}[!tbp]
      \centering      
    \includegraphics[scale=0.4]{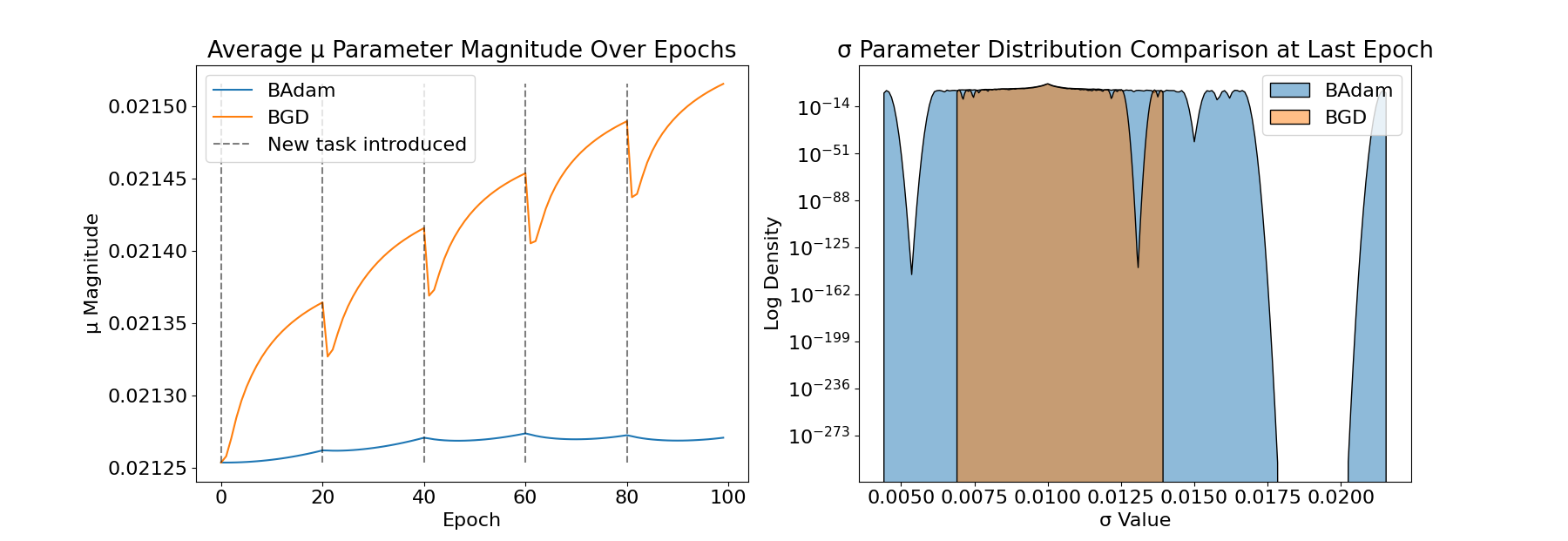}
    \caption{\textbf{Left:} Change in average $\mu$ magnitude during training. Catastrophic forgetting can be observed in BGD's parameters, while this is less present for BAdam.\textbf{ Right:} Distribution of $\sigma$ values at the end of training. BAdam has a wider distribution of uncertainties, indicating that it more effectively constrains highly important parameters while making unused parameters more readily available for learning.}\label{mean_param_value}
   \end{figure*}

\begin{align}    
    &\mathbb{E}_{\epsilon}\left[\frac{\partial L_{n}(\theta)}{\partial\theta_{i}}\epsilon_{i}\right] &&\\ \nonumber
    &= \mathbb{E}_{\epsilon}\left[\left(\frac{\partial L_{n}(\mu)}{\partial\theta_{i}} + \sum_{j} \frac{\partial^{2}L_{n}(\mu)}{\partial \theta_{i} \partial \theta_{j}} \epsilon_{j}\sigma{j} + O(||\sigma^{2}||)    \right)\epsilon_{i}\right]&&\\ \nonumber
    &= \frac{\partial^{2} L_{n}(\mu)}{\partial^{2}\theta_{i}} \sigma_{i} + O(||\sigma||^{2})&&
\end{align}

where for $\epsilon_{i} \text{ and } \epsilon{j}$ the following holds:  $\mathbb{E}_{\epsilon}[\epsilon_{i}] = 0$ and $\mathbb{E}_{\epsilon}[\epsilon_{i}, \epsilon_{j}] = \delta_{ij}$. From equation \ref{std_dev} it is therefore clear that in areas with positive curvature, such as near local and global minima, $\sigma$ will decrease; leading to stronger regularization and better protection of knowledge. Similarly, uncertainty will increase at points of negative curvature, such as near maxima or saddle points. Since BAdam inherits Adam's robustness to saddle points, as well as its faster convergence, BAdam may yield lower values of $\sigma$ than BGD, leading to less forgetting. \\


To motivate these insights empirically, we perform an illustrative experiment to evaluate how parameters change throughout training. Here, we train a feedforward neural network on the SplitMNIST task, initializing $\sigma$ to $0.01$. The first plot in figure \ref{mean_param_value} shows the evolution of the magnitude of $\mu$ for the model trained with BAdam and BGD. From this plot it is clear that BAdam learns smaller parameter values, which is indicative of lower plasticity since parameters can't grow as quickly. Second,parameters optimized by BGD experience a sudden drop in magnitude at the introduction of new tasks, essentially 'soft resetting' the parameters, which is indicative of catastrophic forgetting. In contrast, parameters optimized by BAdam suffer a proportionally smaller drop in magnitude, indicating that these parameters are either better protected, or that they occupy a position in parameter space that better generalizes to subsequent tasks. The second plot of figure \ref{mean_param_value} shows the distribution of $\sigma$ at the end of training. BGD learns values that are fairly close to the initial value of $0.01$, with almost all parameters being closely packed. In contrast, BAdam better disperses uncertainty estimates; this results in many parameters having smaller values for $\sigma$, resulting in better protection, but also a subset of parameters having larger $\sigma$ values, which in turn provides the model with the means to learn downstream tasks.


\section{Experiments}
All experiments are conducted on an Nvidia RTX 4090 and results for continual learning experiments are averaged over 15 seeds $[1,15]$ (25 seeds, for the graduated experiments). We restrict our comparison to popular prior-based approaches that have similar advantages and constraints to BAdam, rather than comparing to paradigms such as memory-based methods, which make different assumptions about the problem domain.\\

\subsection{Cifar10 Convergence Experiment}
In the first experiment, we compare the convergence properties of BAdam to that of BGD and Adam by training a small neural network with 2 convolutional and 3 fully connected layers for 10 epochs on the CIFAR10 dataset. \\

\subsection{Standard Benchmark Experiments}
\begin{table}[!b]
\centering
\caption{SplitMNIST Standard Benchmark Hyper-Parameters}\label{hyper-1}
\begin{center}
\begin{tabular}{|c|c|}
\hline
\textbf{Method} & \textbf{Hyper-Parameters}\\
\hline
\hline
BAdam & $\eta=0.01$ and $\sigma=0.011$\\
\hline
BGD & $\eta=1.0$ and $\sigma=0.01$\\
\hline
MAS & $\alpha=0.5$ and $\lambda=0.1$\\
\hline
EWC & $\lambda=10$ and decay$=0.9$\\
\hline
SI &$\lambda=1.0$\\
\hline
VCL & $\beta=1.0$\\
\hline

\end{tabular}
\end{center}
\end{table}
Next, BAdam's performance is evaluated against other prior-based methods on the three standard benchmark datasets: SplitMNIST, Split FashionMNIST, and PMNIST, as in \citet{farquhar2018towards}. Split MNIST and split FashionMNIST are framed as single-headed class-incremental problems. In split MNIST, the model must successfully classify each digit as in standard MNIST training, however the digits are presented in pairs, sequentially: $[0,1], [2,3], [4,5]$, and so forth. Split FashionMNIST shares the same formulation, but with clothing items. For completeness, we also evaluate on the domain-incremental PMNIST dataset \citep{kirkpatrick2017overcoming}, where a model must classify a sequential set of permuted MNIST digits. We evaluate BAdam against BGD, MAS, EWC, SI, and VCL (without an auxiliary coreset). We adopt the same architectures, hyper-parameters, and experimental setup found in \citet{farquhar2018towards}, with the exception that we do not allow VCL to train for additional epochs on account of being Bayesian, since BGD and BAdam also use Bayesian neural networks but are not afforded additional training time. MAS,  BAdam and BGD hyper-parameters were found via a gridsearch (typically for BAdam we found $\eta$ values in the range $[0.01, 0.1]$ and initial values for $\sigma$ in the range $[0.005, 0.02]$ to be most effective). The implementation utilizes PyTorch \citep{paszke2019pytorch}, and the Avalanche continual learning library \citep{lomonaco2021avalanche}. \\

For both split MNIST and split FashionMNIST, models are trained for 20 epochs (120 for VCL since it is a Bayesian neural network), with a batch size of 256. For split MNIST, a feed-forward neural network with 2 hidden layers of width 256 is trained. For split FashionMNIST, a the network architecture is 4 hidden layers of width 200, and for PMNIST a network with 2 layers of 100 units is trained for 30 epochs.\\



\begin{table}[!t]
\centering
\caption{Split FashionMNIST Standard Benchmark Hyper-Parameters}\label{hyper-2}
\begin{center}
\begin{tabular}{|c|c|}
\hline
\textbf{Method} & \textbf{Hyper-Parameters}\\
\hline
\hline
BAdam &  $\eta=0.01$ and $\sigma=0.005$\\
\hline
BGD & $\eta=1.0$ and $\sigma=0.01$\\
\hline
MAS & $\alpha=0.5$ and $\lambda=0.1$\\
\hline
EWC & $\lambda=10$ and decay$=0.9$\\
\hline
SI &$\lambda=1.0$\\
\hline
VCL & $\beta=1.0$\\
\hline

\end{tabular}
\end{center}
\end{table}


\begin{table}[!h]
\centering
\caption{PMNIST Standard Benchmark Hyper-Parameters}\label{hyper-3}
\begin{center}
\begin{tabular}{|c|c|}
\hline
\textbf{Method} & \textbf{Hyper-Parameters}\\
\hline
\hline
BAdam & $\eta=0.1$ and $\sigma=0.01$\\
\hline
BGD &  $\eta=1.0$ and $\sigma=0.01$\\
\hline
MAS & $\alpha=0.5$ and $\lambda=0.1$\\
\hline
EWC & $\lambda=10$ and decay$=0.9$\\
\hline
SI &$\lambda=1.0$\\
\hline
VCL & $\beta=1.0$\\
\hline

\end{tabular}
\end{center}
\end{table}


\subsection{Graduated Experiments}
\begin{figure}[!bp]\centering
    \includegraphics[scale=0.35]{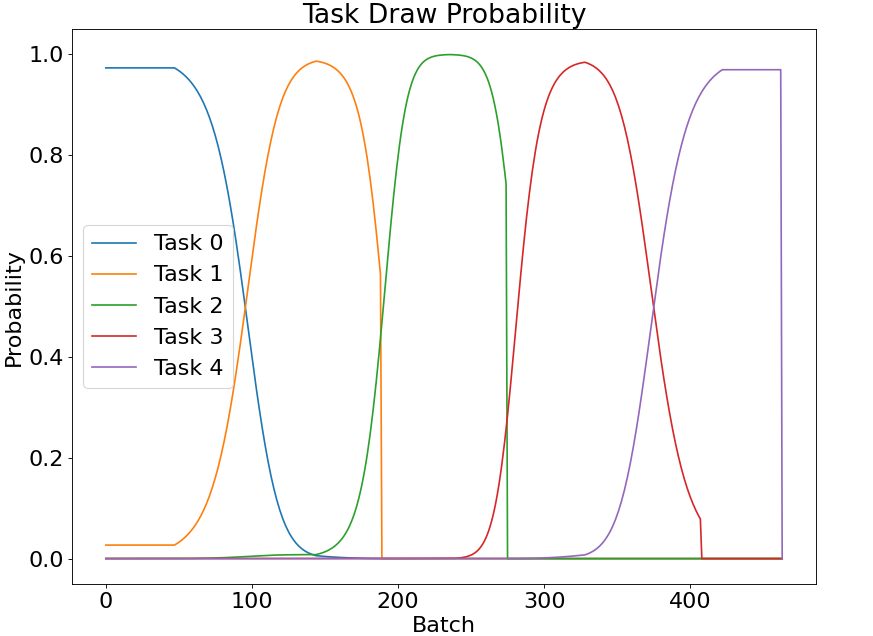}
\caption{Probability of a sample being taken from each task every batch for graduated SplitMNIST}\label{taskprobs}
\end{figure}
We introduce a novel formulation of the standard SplitMNIST benchmarks that features specific conditions reflective of learning in the real world. Referring to these experiments henceforth as the graduated experiments, we take the original SplitMNIST and Split FashionMNIST evaluations and impose the following constraints: first, each data sample is seen only once, since storing and extensively training on data samples is expensive and precludes application on edge devices. Second, task labels are unavailable to the methods, which is reflective of the real-world where an autonomous agent does not necessarily have access to this meta-information. Finally, tasks may have some small overlap, which is representative of how real-world tasks and environments change gradually over time. The graduated boundaries are implemented by assigning draw probabilities to each task based on the index of the current sample. The probability is proportional to the squared distance of the current sample index to the index of the centre point of each task. This creates peaks where a task is near-guaranteed to be selected, which slowly decays into a graduated boundary with the next task. This can be seen in figure \ref{taskprobs}. Since no task labels are available, regularization is recalculated after every batch. While these experiments are similar to those in \citet{zeno2018task}, they are different in that only a single task is permitted and also that we restrict the experiments to single-headed class-incremental problems, which was not enforced in \citet{zeno2018task}. For these tasks, we train a fully-connected neural network with 2 hidden layers of 200 units each, and use the ReLU activation function. We conduct a grid-search for each method's hyper-parameters, optimising for final model performance. In addition to the methods evaluated in the labelled experiments, we also evaluate the Task-Free Continual Learning approach (TFCL), which allows MAS to not require task-labels.\\


      \begin{figure}[!tp]
      \centering      
    \includegraphics[scale=0.35]{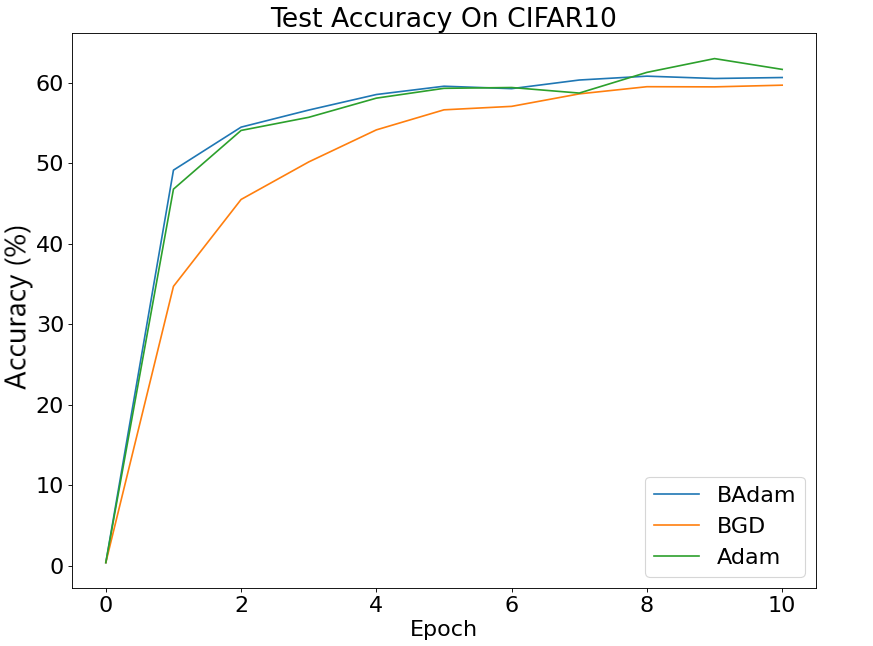}
    \caption{Optimizer convergence rate comparison on single-task CIFAR10. Results show that despite being Bayesian, BAdam can rival the convergence rates of traditional optimizers.}\label{cifar_simple}
   \end{figure}

We train for one epoch (each sample is seen once) with a batch size of 128. A feed-forward neural network with a hidden layer of 200 units is used.

\begin{table}[!h]
\centering
\caption{Graduated Benchmark Hyper-Parameters}\label{hyper-1g}
\begin{center}
\begin{tabular}{|c|c|}
\hline
\textbf{Method} & \textbf{Hyper-Parameters}\\
\hline
\hline
BAdam & $\eta=0.1$ and $\sigma=0.01$\\
\hline
BGD & $\eta=10.0$ and $\sigma=0.017$\\
\hline
MAS & $\alpha=0.6$ and $\lambda=1.0$\\
\hline
EWC & $\lambda=100$ and decay$=0.7$\\
\hline
SI &$\lambda=1.0$\\
\hline
TFCL & $\lambda = 0.4$\\
\hline
VCL & $\beta=0.1$\\
\hline

\end{tabular}
\end{center}
\end{table}



\section{Results}\label{results}
\subsection{Cifar10 Convergence Analysis}

Figure \ref{cifar_simple} shows that BAdam converges much faster than BGD, demonstrating the efficacy of the amended update rule for $\mu$. Furthermore, the convergence rates of Badam are comparable to Adam, which is a strong property since BAdam utilizes a Bayesian neural network. The ability to converge quickly is of significant importance for all optimization tasks, but especially so for domains where compute power is constrained.\\

\subsection{Standard Benchmark Experiments}

\begin{figure}[!tp]\centering
    \includegraphics[scale=0.35]{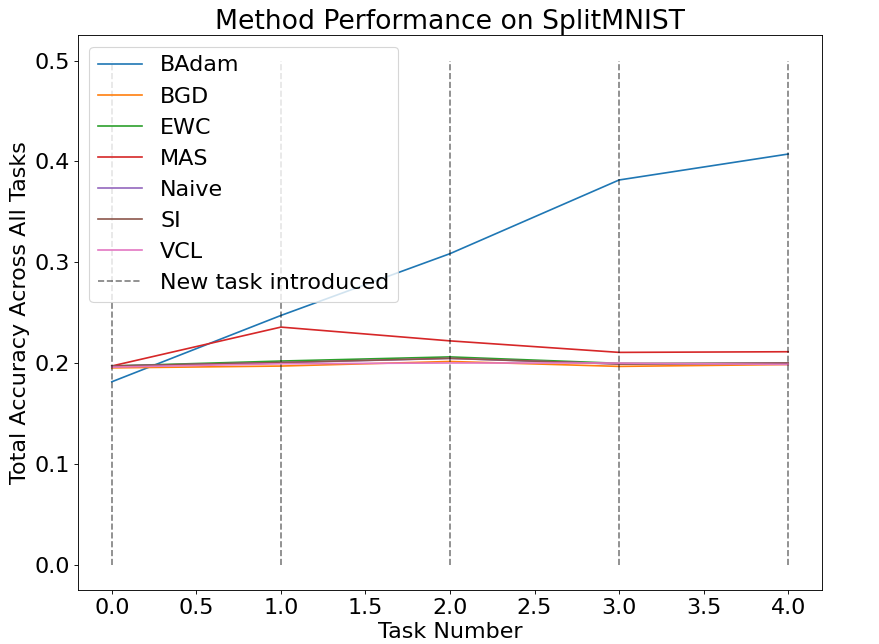}
\caption{CL method performance on class-incremental split MNIST}\label{labelledmnist}
\end{figure}

For split MNIST (figure \ref{labelledmnist}), no previous prior-based method outperforms the na\"ive baseline, all achieving $\sim20\%$ accuracy. In contrast BAdam makes substantial improvements, reaching over $40\%$ accuracy, doubling the efficacy of other methods. On the more challenging Split FashionMNIST benchmark, table \ref{results_table} shows that existing methods once again perform similarly to the na\"ive baseline, achieving $\sim 21\%$ accuracy. BAdam improves upon this significantly, reaching $31\%$ accuracy, being the only method to successfully retain some previous knowledge. These improvements are statistically significant for $p=0.05$ using a standard T-test. While there remains clear room for improvement, BAdam is the first prior based method that makes substantial steps towards solving these challenging single-headed class-incremental tasks. On the domain-incremental PMNIST task, BAdam is competitive with all other methods, whereas VCL's poor performance may be attributed to underfitting.\\
  
\subsection{Graduated Experiments}

\begin{figure}[!bp]\centering
    \includegraphics[scale=0.3]{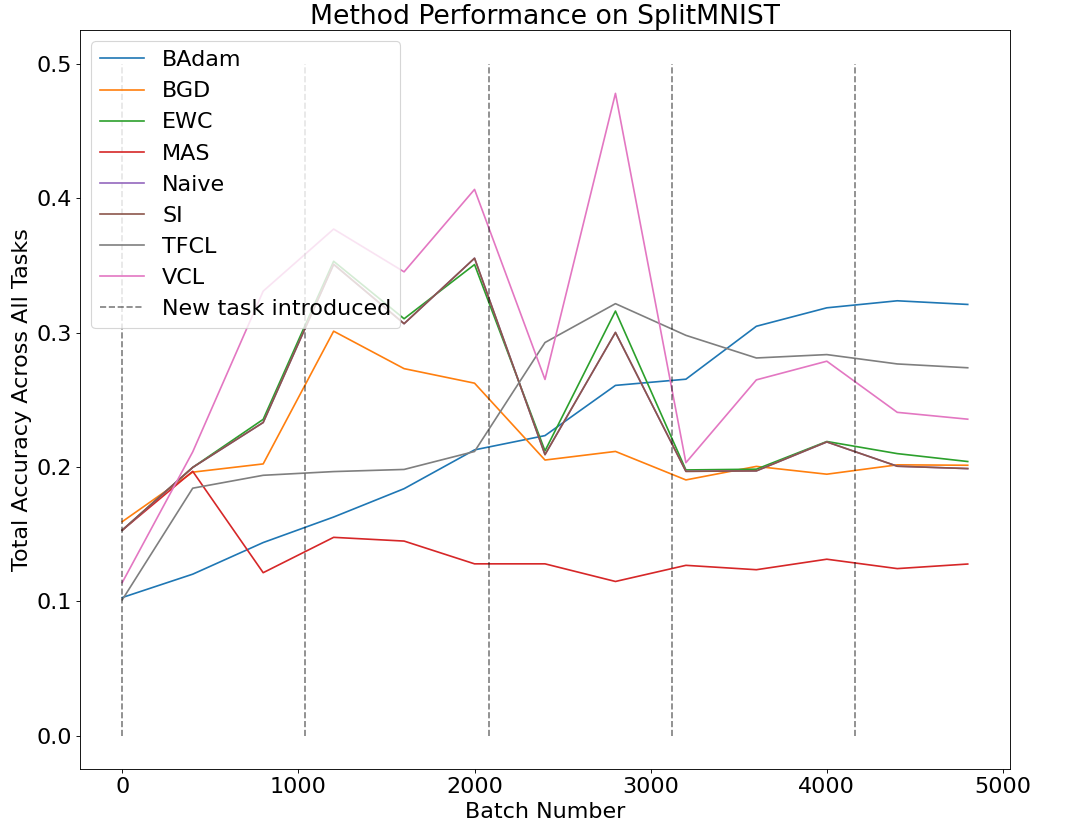}
\caption{CL method performance on class-incremental split MNIST with graduated boundaries, single epoch training, no task labels}\label{gradmnist}
\end{figure}

 As seen in figure \ref{gradmnist} and table \ref{results_table}, BAdam outperforms all other methods on both tasks, and is considerably better than BGD. BAdam not only reaches superior performance as the number of tasks grows, but it is also the only method with stable and consistent improvement. Other methods achieve stronger intermediate performance, but they fail to maintain that level due to catastrophic forgetting. BAdam is seemingly slower to reach good performance compared to other methods, despite its strong convergence properties, however this is explained by method hyper-parameters being optimized for final performance on all tasks, not intermediate performance. BAdam's performance on SplitMNIST is perfectly preserved between the original and graduated experiments, while Split FashionMNIST is slightly worse, which is likey due to underfitting on the more challenging task. These findings indicate that the method is robust to the additional challenges posed by this experiment. Interestingly, VCL improves when applied to the graduated experiments. 

\begin{table*}[!tpb]
\caption{Method Results for All Experiments.}\label{results_table}
\begin{center}
\begin{tabular}{|c|c|c|c|c|c|}
\hline
\textbf{Method} & \textbf{SplitMNIST} & \textbf{Split F-MNIST} & \textbf{PMNIST} & \textbf{Graduated SplitMNIST} & \textbf{Graduated Split F-MNIST}\\
\hline
\hline
Na\"ive & $0.20 \pm{0.02}$  & $0.21 \pm{0.03}$ & $0.65 \pm{0.02}$ & $0.20 \pm{0.01}$ & $0.20 \pm{0.02}$\\
\hline
BAdam & $\boldsymbol{0.41 \pm{0.07}}$ & $\boldsymbol{0.31 \pm{0.07}}$ & $0.81 \pm{0.01}$ & $\boldsymbol{0.41 \pm{0.07}}$ & $\boldsymbol{0.25 \pm{0.07}}$\\
\hline
BGD & $0.20 \pm{0.02}$ & $0.21 \pm{0.03}$ & $0.76 \pm{0.02}$ & $0.20 \pm{0.05}$ & $0.14 \pm{0.05}$\\
\hline
EWC & $0.20 \pm{0.01}$ & $0.21 \pm{0.03}$ & $0.76 \pm{0.01}$ & $0.20 \pm{0.02}$ & $0.21 \pm{0.02}$\\
\hline
MAS & $0.21 \pm{0.03}$ & $0.23 \pm{0.04}$ & $\boldsymbol{0.85 \pm{0.01}}$ & $0.13 \pm{0.03}$ & $0.13 \pm{0.04}$\\
\hline
SI & $0.20 \pm{0.02}$ & $0.21 \pm{0.03}$ & $0.75 \pm{0.02}$ & $0.20 \pm{0.01}$ & $0.20 \pm{0.02}$\\
\hline
VCL & $0.20 \pm{0.01}$ & $0.20 \pm{0.02}$ & $0.51 \pm{0.02}$ & $0.24 \pm{0.07}$ & $0.23 \pm{0.04}$\\
\hline
TFCL & - & - & - & $0.27 \pm{0.07}$ & $0.22 \pm{0.07}$\\
\hline
\end{tabular}
\end{center}
\end{table*}

\section{Discussion}
BAdam exhibits state-of-the-art performance on single-headed, class-incremental tasks for prior-based methods. BAdam is the strongest method evaluated, both in the discrete, labelled domain, and the graduated, label-free, single epoch domain, and is also the first prior-based method to make convincing improvements over the baseline in these experiments. The ability to retain knowledge on these challenging benchmarks is an important step for prior-based methods, although there is still clear room for improvement. \\

Many previous contributions to regularization CL have been new parameter importance estimation methods, however existing prior-based approaches do not seem to have inaccurate importance estimation, and thus further contributions in this direction have failed to improve performance on class-incremental scenarios. The contributions in this work are instead focused around improving the convergence properties of BGD, this seems to be a promising avenue forward.\\

\section{Limitations}

BAdam is promising, however it has two key limitations. Firstly, like BGD, BAdam has two hyper-parameters to optimize, and we found that both of these methods are fairly sensitive to changes in the initialisation value of the standard deviation. Further work on identifying good initialisation values a priori is worthwhile, to help ensure the optimal performance of BAdam may be realised in real-world challenges. Secondly, the primary barrier for progress is the rate at which neural networks can learn more challenging problems. If a model cannot learn a single task in a single epoch (such as complex graph, vision, or reinforcement learning problems), then solving continual learning tasks under such conditions is limited, since the upper bound on performance is unsatisfactory. Future research into fast convergence and concepts such as few-shot learning is important to the progress of online continual learning. Furthermore, since BAdam is a closed-form update rule, if improved learning rates and convergence is reached through advances in stochastic optimization, BAdam will likely need to be further improved with these characteristics. \\

\section{Conclusion}
In this work we present BAdam, which unifies desirable properties of Adam and BGD, yielding a fast-converging continual learning method with no reliance on task labels. The extensions to BGD offered by BAdam are easy to implement, computationally efficient, and built upon familiar concepts from stochastic optimization literature. \\

 We evaluated BAdam alongside a range of regularization CL methods for single-head class-incremental problems in both traditional continual learning conditions, and in a novel experimental setup focused on  continual learning in a single epoch, without requiring task labels or a strongly structured data-stream. This is often more reflective of the real-world and is considerably more challenging than other CL environments \citep{van2019three}. We found BAdam to be the most efficacious approach in both setups, being the only prior-based continual learning method to improve upon the naive baseline and more than doubling the performance of other methods in the SplitMNIST task. While further work is required to fully solve class-incremental problems, BAdam takes the first steps towards retaining previous task knowledge in these challenging domains using prior-based methods, laying the groundwork for important future work in this direction. Further work could explore several avenues, a simple direction would be to explore alternative ways to improve convergence by leveraging other concepts from stochastic optimization. A more interesting direction is to investigate other limitations of prior-based approaches that exist beyond the importance estimation methods.\\








\bibliography{mybibliography}

\begin{thebibliography}{41}
\providecommand{\natexlab}[1]{#1}
\providecommand{\url}[1]{\texttt{#1}}
\expandafter\ifx\csname urlstyle\endcsname\relax
  \providecommand{\doi}[1]{doi: #1}\else
  \providecommand{\doi}{doi: \begingroup \urlstyle{rm}\Url}\fi

\bibitem[Aljundi et~al.(2018)Aljundi, Babiloni, Elhoseiny, Rohrbach, and Tuytelaars]{aljundi2018memory}
Rahaf Aljundi, Francesca Babiloni, Mohamed Elhoseiny, Marcus Rohrbach, and Tinne Tuytelaars.
\newblock Memory aware synapses: Learning what (not) to forget.
\newblock In \emph{Proceedings of the European conference on computer vision (ECCV)}, pages 139--154, 2018.

\bibitem[Aljundi et~al.(2019{\natexlab{a}})Aljundi, Kelchtermans, and Tuytelaars]{Aljundi_2019_CVPR}
Rahaf Aljundi, Klaas Kelchtermans, and Tinne Tuytelaars.
\newblock Task-free continual learning.
\newblock In \emph{Proceedings of the IEEE/CVF Conference on Computer Vision and Pattern Recognition (CVPR)}, June 2019{\natexlab{a}}.

\bibitem[Aljundi et~al.(2019{\natexlab{b}})Aljundi, Lin, Goujaud, and Bengio]{aljundi2019gradient}
Rahaf Aljundi, Min Lin, Baptiste Goujaud, and Yoshua Bengio.
\newblock Gradient based sample selection for online continual learning.
\newblock \emph{Advances in neural information processing systems}, 32, 2019{\natexlab{b}}.

\bibitem[Blundell et~al.(2015)Blundell, Cornebise, Kavukcuoglu, and Wierstra]{blundell2015weight}
Charles Blundell, Julien Cornebise, Koray Kavukcuoglu, and Daan Wierstra.
\newblock Weight uncertainty in neural network.
\newblock In \emph{International conference on machine learning}, pages 1613--1622. PMLR, 2015.

\bibitem[Broderick et~al.(2013)Broderick, Boyd, Wibisono, Wilson, and Jordan]{broderick2013streaming}
Tamara Broderick, Nicholas Boyd, Andre Wibisono, Ashia~C Wilson, and Michael~I Jordan.
\newblock Streaming variational bayes.
\newblock \emph{Advances in neural information processing systems}, 26, 2013.

\bibitem[Castro et~al.(2018)Castro, Mar{\'\i}n-Jim{\'e}nez, Guil, Schmid, and Alahari]{castro2018end}
Francisco~M Castro, Manuel~J Mar{\'\i}n-Jim{\'e}nez, Nicol{\'a}s Guil, Cordelia Schmid, and Karteek Alahari.
\newblock End-to-end incremental learning.
\newblock In \emph{Proceedings of the European conference on computer vision (ECCV)}, pages 233--248, 2018.

\bibitem[Cichon and Gan(2015)]{cichon2015branch}
Joseph Cichon and Wen-Biao Gan.
\newblock Branch-specific dendritic ca2+ spikes cause persistent synaptic plasticity.
\newblock \emph{Nature}, 520\penalty0 (7546):\penalty0 180--185, 2015.

\bibitem[Cohen et~al.(1997)Cohen, Celnik, Pascual-Leone, Corwell, Faiz, Dambrosia, Honda, Sadato, Gerloff, Hallett, et~al.]{cohen1997functional}
Leonardo~G Cohen, Pablo Celnik, Alvaro Pascual-Leone, Brian Corwell, Lala Faiz, James Dambrosia, Manabu Honda, Norihiro Sadato, Christian Gerloff, Mark Hallett, et~al.
\newblock Functional relevance of cross-modal plasticity in blind humans.
\newblock \emph{Nature}, 389\penalty0 (6647):\penalty0 180--183, 1997.

\bibitem[Draelos et~al.(2017)Draelos, Miner, Lamb, Cox, Vineyard, Carlson, Severa, James, and Aimone]{draelos2017neurogenesis}
Timothy~J Draelos, Nadine~E Miner, Christopher~C Lamb, Jonathan~A Cox, Craig~M Vineyard, Kristofor~D Carlson, William~M Severa, Conrad~D James, and James~B Aimone.
\newblock Neurogenesis deep learning: Extending deep networks to accommodate new classes.
\newblock In \emph{2017 International Joint Conference on Neural Networks (IJCNN)}, pages 526--533. IEEE, 2017.

\bibitem[Ebrahimi et~al.(2019)Ebrahimi, Elhoseiny, Darrell, and Rohrbach]{ebrahimi2019uncertainty}
Sayna Ebrahimi, Mohamed Elhoseiny, Trevor Darrell, and Marcus Rohrbach.
\newblock Uncertainty-guided continual learning with bayesian neural networks.
\newblock \emph{arXiv preprint arXiv:1906.02425}, 2019.

\bibitem[Farquhar and Gal(2018)]{farquhar2018towards}
Sebastian Farquhar and Yarin Gal.
\newblock Towards robust evaluations of continual learning.
\newblock \emph{arXiv preprint arXiv:1805.09733}, 2018.

\bibitem[French(1999)]{french1999catastrophic}
Robert~M French.
\newblock Catastrophic forgetting in connectionist networks.
\newblock \emph{Trends in cognitive sciences}, 3\penalty0 (4):\penalty0 128--135, 1999.

\bibitem[Graves(2011)]{NIPS2011_7eb3c8be}
Alex Graves.
\newblock Practical variational inference for neural networks.
\newblock In J.~Shawe-Taylor, R.~Zemel, P.~Bartlett, F.~Pereira, and K.Q. Weinberger, editors, \emph{Advances in Neural Information Processing Systems}, volume~24. Curran Associates, Inc., 2011.

\bibitem[Hassabis et~al.(2017)Hassabis, Kumaran, Summerfield, and Botvinick]{hassabis2017neuroscience}
Demis Hassabis, Dharshan Kumaran, Christopher Summerfield, and Matthew Botvinick.
\newblock Neuroscience-inspired artificial intelligence.
\newblock \emph{Neuron}, 95\penalty0 (2):\penalty0 245--258, 2017.

\bibitem[Hecht-Nielsen(1992)]{hecht1992theory}
Robert Hecht-Nielsen.
\newblock Theory of the backpropagation neural network.
\newblock In \emph{Neural networks for perception}, pages 65--93. Elsevier, 1992.

\bibitem[Jospin et~al.(2022)Jospin, Laga, Boussaid, Buntine, and Bennamoun]{jospin2022hands}
Laurent~Valentin Jospin, Hamid Laga, Farid Boussaid, Wray Buntine, and Mohammed Bennamoun.
\newblock Hands-on bayesian neural networks—a tutorial for deep learning users.
\newblock \emph{IEEE Computational Intelligence Magazine}, 17\penalty0 (2):\penalty0 29--48, 2022.

\bibitem[Khan et~al.(2018)Khan, Nielsen, Tangkaratt, Lin, Gal, and Srivastava]{khan2018fast}
Mohammad Khan, Didrik Nielsen, Voot Tangkaratt, Wu~Lin, Yarin Gal, and Akash Srivastava.
\newblock Fast and scalable bayesian deep learning by weight-perturbation in adam.
\newblock In \emph{International conference on machine learning}, pages 2611--2620. PMLR, 2018.

\bibitem[Kingma and Ba(2014)]{kingma2014adam}
Diederik~P Kingma and Jimmy Ba.
\newblock Adam: A method for stochastic optimization.
\newblock \emph{arXiv preprint arXiv:1412.6980}, 2014.

\bibitem[Kirkpatrick et~al.(2017)Kirkpatrick, Pascanu, Rabinowitz, Veness, Desjardins, Rusu, Milan, Quan, Ramalho, Grabska-Barwinska, et~al.]{kirkpatrick2017overcoming}
James Kirkpatrick, Razvan Pascanu, Neil Rabinowitz, Joel Veness, Guillaume Desjardins, Andrei~A Rusu, Kieran Milan, John Quan, Tiago Ramalho, Agnieszka Grabska-Barwinska, et~al.
\newblock Overcoming catastrophic forgetting in neural networks.
\newblock \emph{Proceedings of the national academy of sciences}, 114\penalty0 (13):\penalty0 3521--3526, 2017.

\bibitem[Li and Hoiem(2017)]{li2017learning}
Zhizhong Li and Derek Hoiem.
\newblock Learning without forgetting.
\newblock \emph{IEEE transactions on pattern analysis and machine intelligence}, 40\penalty0 (12):\penalty0 2935--2947, 2017.

\bibitem[Lomonaco et~al.(2021)Lomonaco, Pellegrini, Cossu, Carta, Graffieti, Hayes, Lange, Masana, Pomponi, van~de Ven, Mundt, She, Cooper, Forest, Belouadah, Calderara, Parisi, Cuzzolin, Tolias, Scardapane, Antiga, Amhad, Popescu, Kanan, van~de Weijer, Tuytelaars, Bacciu, and Maltoni]{lomonaco2021avalanche}
Vincenzo Lomonaco, Lorenzo Pellegrini, Andrea Cossu, Antonio Carta, Gabriele Graffieti, Tyler~L. Hayes, Matthias~De Lange, Marc Masana, Jary Pomponi, Gido van~de Ven, Martin Mundt, Qi~She, Keiland Cooper, Jeremy Forest, Eden Belouadah, Simone Calderara, German~I. Parisi, Fabio Cuzzolin, Andreas Tolias, Simone Scardapane, Luca Antiga, Subutai Amhad, Adrian Popescu, Christopher Kanan, Joost van~de Weijer, Tinne Tuytelaars, Davide Bacciu, and Davide Maltoni.
\newblock Avalanche: an end-to-end library for continual learning.
\newblock In \emph{Proceedings of IEEE Conference on Computer Vision and Pattern Recognition}, 2nd Continual Learning in Computer Vision Workshop, 2021.

\bibitem[Lopez-Paz and Ranzato(2017)]{lopez2017gradient}
David Lopez-Paz and Marc'Aurelio Ranzato.
\newblock Gradient episodic memory for continual learning.
\newblock \emph{Advances in neural information processing systems}, 30, 2017.

\bibitem[McCloskey and Cohen(1989)]{mccloskey1989catastrophic}
Michael McCloskey and Neal~J Cohen.
\newblock Catastrophic interference in connectionist networks: The sequential learning problem.
\newblock In \emph{Psychology of learning and motivation}, volume~24, pages 109--165. Elsevier, 1989.

\bibitem[Nguyen et~al.(2017)Nguyen, Li, Bui, and Turner]{nguyen2017variational}
Cuong~V Nguyen, Yingzhen Li, Thang~D Bui, and Richard~E Turner.
\newblock Variational continual learning.
\newblock \emph{arXiv preprint arXiv:1710.10628}, 2017.

\bibitem[Parisi et~al.(2019)Parisi, Kemker, Part, Kanan, and Wermter]{parisi2019continual}
German~I Parisi, Ronald Kemker, Jose~L Part, Christopher Kanan, and Stefan Wermter.
\newblock Continual lifelong learning with neural networks: A review.
\newblock \emph{Neural networks}, 113:\penalty0 54--71, 2019.

\bibitem[Paszke et~al.(2019)Paszke, Gross, Massa, Lerer, Bradbury, Chanan, Killeen, Lin, Gimelshein, Antiga, et~al.]{paszke2019pytorch}
Adam Paszke, Sam Gross, Francisco Massa, Adam Lerer, James Bradbury, Gregory Chanan, Trevor Killeen, Zeming Lin, Natalia Gimelshein, Luca Antiga, et~al.
\newblock Pytorch: An imperative style, high-performance deep learning library.
\newblock \emph{Advances in neural information processing systems}, 32, 2019.

\bibitem[Rannen et~al.(2017)Rannen, Aljundi, Blaschko, and Tuytelaars]{rannen2017encoder}
Amal Rannen, Rahaf Aljundi, Matthew~B Blaschko, and Tinne Tuytelaars.
\newblock Encoder based lifelong learning.
\newblock In \emph{Proceedings of the IEEE International Conference on Computer Vision}, pages 1320--1328, 2017.

\bibitem[Rebuffi et~al.(2017)Rebuffi, Kolesnikov, Sperl, and Lampert]{rebuffi2017icarl}
Sylvestre-Alvise Rebuffi, Alexander Kolesnikov, Georg Sperl, and Christoph~H Lampert.
\newblock icarl: Incremental classifier and representation learning.
\newblock In \emph{Proceedings of the IEEE conference on Computer Vision and Pattern Recognition}, pages 2001--2010, 2017.

\bibitem[Rolnick et~al.(2019)Rolnick, Ahuja, Schwarz, Lillicrap, and Wayne]{rolnick2019experience}
David Rolnick, Arun Ahuja, Jonathan Schwarz, Timothy Lillicrap, and Gregory Wayne.
\newblock Experience replay for continual learning.
\newblock \emph{Advances in Neural Information Processing Systems}, 32, 2019.

\bibitem[Rusu et~al.(2016)Rusu, Rabinowitz, Desjardins, Soyer, Kirkpatrick, Kavukcuoglu, Pascanu, and Hadsell]{rusu2016progressive}
Andrei~A Rusu, Neil~C Rabinowitz, Guillaume Desjardins, Hubert Soyer, James Kirkpatrick, Koray Kavukcuoglu, Razvan Pascanu, and Raia Hadsell.
\newblock Progressive neural networks.
\newblock \emph{arXiv preprint arXiv:1606.04671}, 2016.

\bibitem[Shin et~al.(2017)Shin, Lee, Kim, and Kim]{shin2017continual}
Hanul Shin, Jung~Kwon Lee, Jaehong Kim, and Jiwon Kim.
\newblock Continual learning with deep generative replay.
\newblock \emph{Advances in neural information processing systems}, 30, 2017.

\bibitem[Sussmann(1992)]{sussmann1992uniqueness}
H{\'e}ctor~J Sussmann.
\newblock Uniqueness of the weights for minimal feedforward nets with a given input-output map.
\newblock \emph{Neural networks}, 5\penalty0 (4):\penalty0 589--593, 1992.

\bibitem[Terekhov et~al.(2015)Terekhov, Montone, and O’Regan]{terekhov2015knowledge}
Alexander~V Terekhov, Guglielmo Montone, and J~Kevin O’Regan.
\newblock Knowledge transfer in deep block-modular neural networks.
\newblock In \emph{Conference on Biomimetic and Biohybrid Systems}, pages 268--279. Springer, 2015.

\bibitem[Thrun and Mitchell(1995)]{THRUN199525}
Sebastian Thrun and Tom~M. Mitchell.
\newblock Lifelong robot learning.
\newblock \emph{Robotics and Autonomous Systems}, 15\penalty0 (1):\penalty0 25--46, 1995.
\newblock ISSN 0921-8890.
\newblock \doi{https://doi.org/10.1016/0921-8890(95)00004-Y}.
\newblock URL \url{https://www.sciencedirect.com/science/article/pii/092188909500004Y}.
\newblock The Biology and Technology of Intelligent Autonomous Agents.

\bibitem[Titsias et~al.(2020)Titsias, Schwarz, Matthews, Pascanu, and Teh]{titsias2019functional}
Michalis~K Titsias, Jonathan Schwarz, Alexander G de~G Matthews, Razvan Pascanu, and Yee~Whye Teh.
\newblock Functional regularisation for continual learning using gaussian processes.
\newblock \emph{International Conference on Learning Representations}, 2020.

\bibitem[Van~de Ven and Tolias(2019)]{van2019three}
Gido~M Van~de Ven and Andreas~S Tolias.
\newblock Three scenarios for continual learning.
\newblock \emph{arXiv preprint arXiv:1904.07734}, 2019.

\bibitem[Wang et~al.(2022{\natexlab{a}})Wang, Zhang, Ebrahimi, Sun, Zhang, Lee, Ren, Su, Perot, Dy, et~al.]{wang2022dualprompt}
Zifeng Wang, Zizhao Zhang, Sayna Ebrahimi, Ruoxi Sun, Han Zhang, Chen-Yu Lee, Xiaoqi Ren, Guolong Su, Vincent Perot, Jennifer Dy, et~al.
\newblock Dualprompt: Complementary prompting for rehearsal-free continual learning.
\newblock In \emph{European Conference on Computer Vision}, pages 631--648. Springer, 2022{\natexlab{a}}.

\bibitem[Wang et~al.(2022{\natexlab{b}})Wang, Zhang, Lee, Zhang, Sun, Ren, Su, Perot, Dy, and Pfister]{wang2022learning}
Zifeng Wang, Zizhao Zhang, Chen-Yu Lee, Han Zhang, Ruoxi Sun, Xiaoqi Ren, Guolong Su, Vincent Perot, Jennifer Dy, and Tomas Pfister.
\newblock Learning to prompt for continual learning.
\newblock In \emph{Proceedings of the IEEE/CVF Conference on Computer Vision and Pattern Recognition}, pages 139--149, 2022{\natexlab{b}}.

\bibitem[Wo{\l}czyk et~al.(2022)Wo{\l}czyk, Piczak, W{\'o}jcik, Pustelnik, Morawiecki, Tabor, Trzcinski, and Spurek]{wolczyk2022continual}
Maciej Wo{\l}czyk, Karol Piczak, Bartosz W{\'o}jcik, Lukasz Pustelnik, Pawe{\l} Morawiecki, Jacek Tabor, Tomasz Trzcinski, and Przemys{\l}aw Spurek.
\newblock Continual learning with guarantees via weight interval constraints.
\newblock In \emph{International Conference on Machine Learning}, pages 23897--23911. PMLR, 2022.

\bibitem[Zenke et~al.(2017)Zenke, Poole, and Ganguli]{pmlr-v70-zenke17a}
Friedemann Zenke, Ben Poole, and Surya Ganguli.
\newblock Continual learning through synaptic intelligence.
\newblock In Doina Precup and Yee~Whye Teh, editors, \emph{Proceedings of the 34th International Conference on Machine Learning}, volume~70 of \emph{Proceedings of Machine Learning Research}, pages 3987--3995. PMLR, 06--11 Aug 2017.
\newblock URL \url{https://proceedings.mlr.press/v70/zenke17a.html}.

\bibitem[Zeno et~al.(2018)Zeno, Golan, Hoffer, and Soudry]{zeno2018task}
Chen Zeno, Itay Golan, Elad Hoffer, and Daniel Soudry.
\newblock Task agnostic continual learning using online variational bayes.
\newblock \emph{arXiv preprint arXiv:1803.10123}, 2018.

\end{thebibliography}

\newpage

\onecolumn

\end{document}